\documentclass[letterpaper, 10 pt, conference]{ieeeconf} 
\IEEEoverridecommandlockouts    

\makeatletter
\def\ps@IEEEtitlepagestyle{%
  \def\@oddfoot{%
    \parbox{\textwidth}{\scriptsize
  
    Copyright © 2025 IEEE. All rights reserved. This is the accepted version of the paper. The final version will be published in the \textit{IEEE International Conference on Intelligent Transportation Systems (ITSC), 2025}}
    }%
  }%

\makeatother

\usepackage{multirow}
\usepackage{lipsum}
\usepackage{amsmath}
\usepackage{amssymb}
\usepackage{graphicx}
\graphicspath{{./Figures/}}
\usepackage{algorithm}
\usepackage{multirow}

\usepackage{algpseudocode}
\newtheorem{thm}{Problem}
\usepackage{eso-pic}

\usepackage{xcolor}

\usepackage{amsmath,amsfonts,bm}
\usepackage{xspace}

\newcommand{\bmu}{{\boldsymbol{\mu}}}

\newcommand{\model}{TrafficDiffuser}

\def\eqref#1{equation~\ref{#1}}

\def\1{\bm{1}}

\def\rvh{{\mathbf{h}}}

\def\rvs{{\mathbf{s}}}

\DeclareMathAlphabet{\mathsfit}{\encodingdefault}{\sfdefault}{m}{sl}
\SetMathAlphabet{\mathsfit}{bold}{\encodingdefault}{\sfdefault}{bx}{n}

\usepackage[acronym]{glossaries}
\newacronym{SGD}{\textsc{sgd}}{stochastic gradient descent}
\newacronym{MAP}{\textsc{map}}{maximum-a-posteriori}
\newacronym{MLE}{\textsc{mle}}{maximum likelihood estimation}
\newacronym{MNLL}{\textsc{mnll}}{mean negative log-likelihood}
\newacronym{NLL}{\textsc{nll}}{negative log-likelihood}
\newacronym{LL}{\textsc{ll}}{log-likelihood}
\newacronym{RMSE}{\textsc{rmse}}{root mean square error}
\newacronym{ECE}{\textsc{ece}}{expected calibration error}
\newacronym{SNR}{\textsc{snr}}{signal-to-noise ratio}
\newacronym{FID}{\textsc{fid}}{Fr\'echet Inception Distance}
\newacronym{BPD}{\textsc{bpd}}{bit per dimension}
\newacronym{NFE}{\textsc{nfe}}{neural function evaluations}

\newacronym{AE}{\textsc{ae}}{auto-encoder}
\newacronym{WAE}{\textsc{wae}}{Wasserstein Auto-encoder}
\newacronym{VAE}{\textsc{vae}}{Variational Auto-encoder}
\newacronym{BAE}{\textsc{bae}}{Bayesian Auto-encoder}
\newacronym{CDF}{\textsc{cdf}}{cumulative density function}
\newacronym{GAN}{\textsc{gan}}{Generative Adversarial Network}
\newacronym{DPGMM}{\textsc{dpgmm}}{Dirichlet process Gaussian mixture model}
\newacronym{GMM}{\textsc{gmm}}{Gaussian mixture model}

\newacronym{MC}{mc}{Monte Carlo}
\newacronym{SDE}{\textsc{sde}}{Stochastic Differential Equation}
\newacronym{CNF}{cnf}{Continuous Normaxlizing Flow}
\newacronym{ODE}{ode}{Ordinary Differential Equation}
\newacronym{MCMC}{\textsc{mcmc}}{Markov chain Monte Carlo}
\newacronym{HMC}{\textsc{hmc}}{Hamiltonian Monte Carlo}
\newacronym{MH}{mh}{Metropolis-Hastings}
\newacronym{NUTS}{nuts}{no-u-turn sampler}
\newacronym{SGHMC}{\textsc{sghmc}}{stochastic gradient Hamiltonian Monte Carlo}

\newacronym[longplural=deep Gaussian processes]{DGP}{\textsc{dgp}}{deep Gaussian process} 
\newacronym{GPLVM}{gplvm}{Gaussian process latent variable model}
\newacronym{DPMM}{dpmm}{Dirichlet Process Mixture Model}

\newacronym{VFE}{vfe}{variational free energy}

\newacronym[longplural=Gaussian Processes]{GP}{\textsc{gp}}{Gaussian Process}

\newacronym{VI}{\textsc{vi}}{variational inference}
\newacronym{SVI}{\textsc{svi}}{stochastic variational inference}

\newacronym{ELBO}{\textsc{elbo}}{evidence lower bound}
\newacronym{NELBO}{\textsc{nelbo}}{negative evidence lower bound}
\newacronym{ELL}{\textsc{ell}}{expected log likelihood}
\newacronym{KL}{\textsc{kl}}{Kullback-Leibler}
\newacronym{AUC}{auc}{area under the curve}

\newacronym{BNN}{\textsc{bnn}}{Bayesian neural network}
\newacronym{DNN}{\textsc{dnn}}{deep neural network}
\newacronym{CNN}{\textsc{cnn}}{convolutional neural network}
\newacronym{MLP}{\textsc{mlp}}{multilayer perceptron}
\newacronym{NN}{nn}{neural network}
\newacronym{RELU}{ReLU}{rectified linear unit}

\newacronym{NF}{nf}{normalizing flow}

\newacronym{RBF}{rbf}{radial basis function}
\newacronym{ARD}{ard}{automatic relevance determination}

\newacronym{RKHS}{rkhs}{reproducing kernel Hilbert space}

\newacronym{OT}{ot}{optimal transport}
\newacronym{WD}{wd}{Wasserstein distance}
\newacronym{SWD}{swd}{sliced-Wasserstein distance}
\newacronym{DSWD}{dswd}{distributional sliced-Wasserstein distance}
\newacronym{fsp}{FSP}{Fictitious Self Play}
\newacronym{marl}{MARL}{Multi-agent reinforcement learning}
\newacronym{pomg}{POMG}{Partially Observable Markov Games}
\newacronym{ddpm}{DDPM}{Denoising Diffusion Probabilistic Models}
\newacronym{rl}{RL}{Reinforcement Learning}
\newacronym{sl}{SL}{Supervised Learning}
\newacronym{dp}{DP}{Diffusion Policy}

\newacronym{mpe}{MPE}{Multiple Particle Environment}
\newacronym{pd}{PD}{{Path Diffuser}}
\newacronym{pd_p}{PD$\ominus$P}{PD without Primitives}
\newacronym{dt}{DIFFT}{Differential Transformer}
\newacronym{ff}{Frenet Frame}{The Frenet-Serret Frame}
\newacronym{mha}{MHA}{Multi-Head Attention}
\newacronym{hmp}{HMP}{Heterogeneous Message Passing}
\newacronym{vd}{VD}{Vanilla Diffusion}
\newacronym{cdb}{CDB}{Centralized and Decentralized Behavior}

\usepackage[hidelinks]{hyperref}

\title{\LARGE \bf Top-down Traffic Scenario Generation via Joint Initial-Goal Diffusion and Trajectory Infilling}

\author{
   Da Saem Lee, Yash Vardhan Pant, Sebastian Fischmeister
   \thanks{This work is supported in part by the Natural Sciences and Engineering
Research Council of Canada (NSERC), Canada Foundation for Innovation - John R. Evans Leaders Fund (CFI JELF), Mitacs, and Intact Financial Corporation.}
   \thanks{The authors are with the Department of Electrical and Computer Engineering, University of Waterloo, Waterloo, Canada.
 \ttfamily{ds3lee@uwaterloo.ca, yash.pant@uwaterloo.ca, sfischme@uwaterloo.ca}
}
}

\begin{document}

\bstctlcite{IEEEexample:BSTcontrol} 

\maketitle
\AddToShipoutPictureFG*{%
  \AtPageLowerLeft{%
    \raisebox{6mm}{%
      \makebox[\paperwidth][c]{%
        \parbox{0.9\paperwidth}{%
          \centering\scriptsize
        Copyright \textcopyright\ 2026 IEEE. All rights reserved. This is the accepted version of the paper.The final version will be published \\ in the \textit{IEEE International Conference on Intelligent Transportation Systems (ITSC), 2026}
        }%
      }%
    }%
  }%
}

\begin{abstract}
Robust traffic simulators are crucial for developing and testing autonomous vehicles to reduce the costly, labor-intensive real-world data collection process and the need for physical presence on the road.
However, existing simulators require agents' initial states to generate trajectories, which limits scalability and diversity due to restrictions on the given initial states. While data-driven agent initialization has been widely studied, the generated initial states are not interpretable in terms of why the agents are initialized at those specific locations. Given known initial states, trajectory generation is also a challenging problem, as the model must learn the variability of the destination and how agents should reach it over time.
In this paper, we propose \model{}, a top-down traffic scenario generation framework that generates high-level traffic scenarios, defined by initial and goal state pairs, by jointly modeling them. The high-level scenario generation makes initial states better interpretable and reduces trajectory generation into as simple as an infilling problem.
We demonstrate how the generated high-level traffic scenarios can be used, including constraining based on different trajectory modes and integrating them with existing trajectory generation models. We conduct extensive experiments on the Argoverse 2 motion prediction dataset to evaluate how well the generated outputs capture real-world distributions. 
In addition to generating goal states, \model{} outperforms the next-best approach for agent initialization, reducing speed distribution distance by 55.3\% and the off-road rate by 2.8\%. \footnote{\url{https://github.com/CL2-UWaterloo/TrafficDiffuser}}
\end{abstract}

\section{Introduction}
When developing the autonomous system, exposure to diverse traffic scenarios is crucial to ensure safety and adaptability. However, obtaining and manually annotating real-world datasets that capture such diversity is both time-consuming and costly~\cite{wilson2023argoverse2generationdatasets}. To mitigate this, data-driven simulators~\cite{jiang2024scenediffuserefficientcontrollabledriving, rowe2025scenariodreamer} focuses on generating synthetic traffic scenarios to create data samples. Typically, this process is decoupled into two tasks: generating initial agent states~\cite{lu2024scenecontrol, Tan_2021_CVPR} and then synthesizing the corresponding trajectories~\cite{jiang2024scenediffuserefficientcontrollabledriving}.

For multi-agent scenarios, generating a realistic set of initial agent states is important for diversifying the scenario. While prior data-driven approaches~\cite{lu2024scenecontrol, Tan_2021_CVPR} model the initial states and interactions conditioned on the map, they often lack interpretability, as it is difficult to understand the underlying intent of the generated states. By treating initialization as a task to capture a single timestep of scenarios, these models provide limited context to trajectory generation.

With the given initial traffic scene, trajectory generation has been studied to complete the scenario generation process~\cite{jiang2024scenediffuserefficientcontrollabledriving, rowe2024ctrlsimreactivecontrollabledriving}. Effective trajectory generation requires the model to capture agents' intentions and behaviors. In trajectory prediction tasks, given the initial states and past trajectories, prior works~\cite{grimm2025goalbasedtrajectorypredictionimproved, zhao2020tnttargetdriventrajectoryprediction} focus on predicting goal points and motions between the initial and predicted goal points to improve prediction accuracy in unseen map structures. 

\begin{figure}[t]
        \centering
    \includegraphics[width=0.9\linewidth]{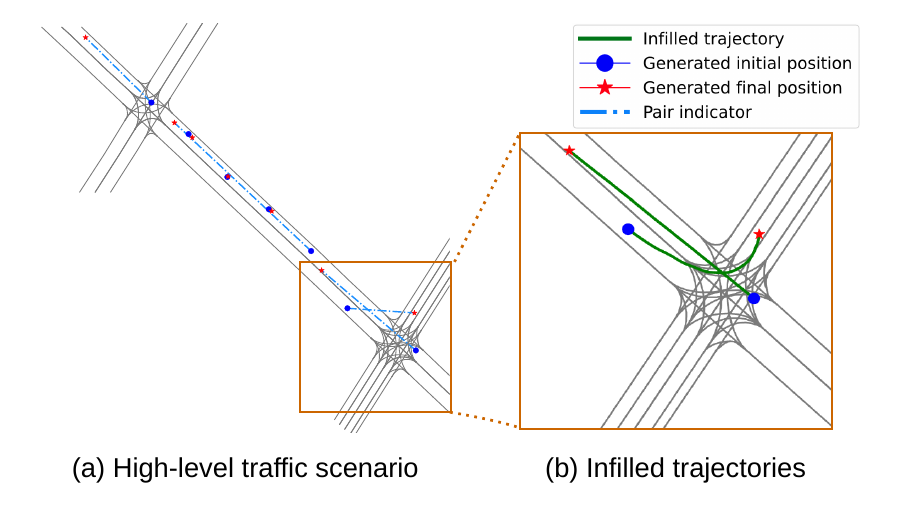}
    \vspace{-5mm}
    \caption{Illustration of proposed approach. (a) Generated high-level traffic scenario. (b) Infilled trajectories conditioned on the high-level scenario.}
    \label{fig:teasor}
    \vspace{-8mm}
\end{figure}
In this paper, we introduce \model{}, a diffusion-based framework that generates traffic scenarios in a top-down, goal-conditioned manner. As illustrated in  Fig.~\ref{fig:teasor}(a), the initial and goal states are jointly modeled to generate high-level scenarios which allows inherently learning feasible paths between endpoints. This high-level scenario simplifies the complex trajectory generation into a goal-conditioned infilling task as shown in Fig.~\ref{fig:teasor}(b). Fig.~\ref{fig:overview} provides an overview of the proposed framework, which integrates high-level scenario generation with trajectory infilling. The contributions of our approaches are as follows:

\begin{figure*}[tb]
    \centering
    \includegraphics[width=0.95\linewidth]{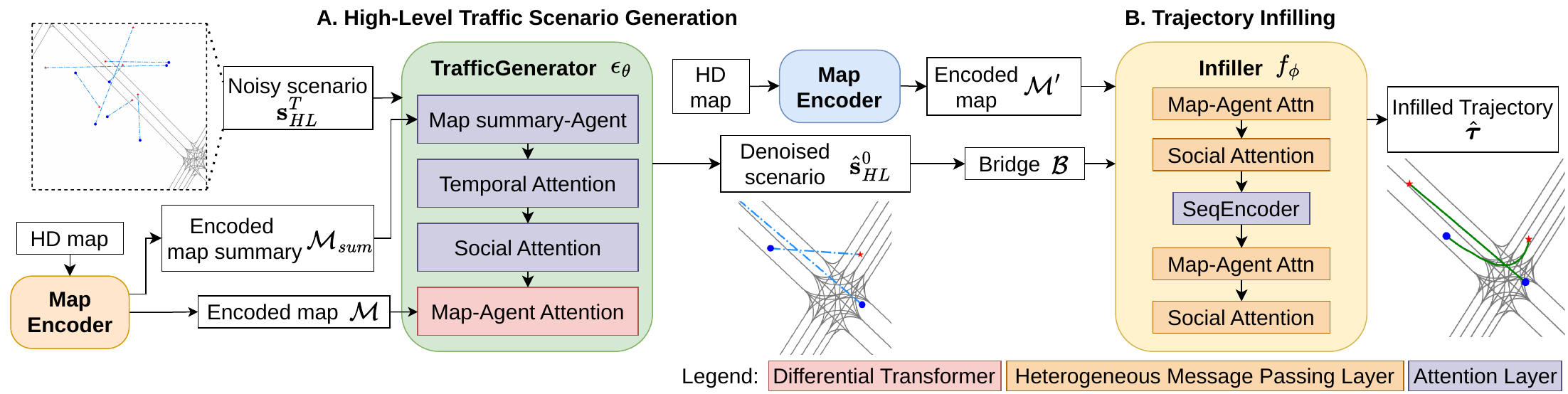}
    \vspace{-3mm}
    \caption{Overview of \model{}. In \textbf{A}, a noisy scenario state, map summary, and map are processed by the  TrafficGenerator $\epsilon_\theta$ to produce a denoised high-level scenario $\hat{\rvs}^0$, which consists of initial and goal states. In \textbf{B}, the denoised scenario is then passed to infiller $f_\phi$ to connect between the generated initial and goal states with kinematically feasible trajectories $\hat{\bm\tau}$.}
    \label{fig:overview}
    \vspace{-5mm}
\end{figure*}

\begin{itemize}
    \item \textbf{Interpretability}: Jointly generated start-goal pair provides better interpretability of the initial traffic scenes and agent intentions.

    \item \textbf{Feasible path generation}: The model synthesizes feasible start-goal configurations for multi-agent scenarios.
    \item \textbf{Simplified Trajectory Generation}: 
    Leveraging high-level traffic scenarios enables simplified trajectory generation through goal-conditioned infilling.

\end{itemize}

The remainder of the paper is organized as follows. Section~\ref{sec:relatedworks} reviews prior works, Section~\ref{sec:preliminaries} introduces notation, problem setup, and preliminaries, Section~\ref{sec:approach} details our approach, and Section~\ref{sec:experiments} presents experimental results and comparisons. Finally, Section~\ref{sec:conclusion} concludes the paper and discusses future work.
\vspace{-2mm}

\section{Related Works}
\label{sec:relatedworks}
\vspace{-1mm}

In this section, we review existing works for traffic scenario generation. We specifically focus on data-driven approaches that initialize agents in the scene, predict or generate their future trajectories, and end-to-end approaches.

Earlier works~\cite{Chitta2024ECCV, sun2024drivescenegen} represented traffic scenarios using rasterized images, by encoding individual features in each channel. To improve computational efficiency by eliminating redundancy, recent works employ vectorized representations of the map structure~\cite{liang2020learninglanegraphrepresentations, gao2020vectornetencodinghdmaps}. SceneControl~\cite{lu2024scenecontrol} leverages diffusion models with transformer decoder layers to generate initial states of the agents in the scene. While SceneControl demonstrates controllable agent initialization on diverse scenarios, it requires guidance sampling~\cite{dhariwal2021diffusion} to enforce realism, such as collision avoidance and road compliance.

With known initial states, trajectory prediction or generation tasks are widely studied using vectorized map representations. For trajectory generation, Ctrl-Sim~\cite{rowe2024ctrlsimreactivecontrollabledriving} introduces a simulator trained using offline reinforcement learning (RL). The real-world driving logs are replayed on the Nocturne simulator~\cite{vinitsky2023nocturnescalabledrivingbenchmark} to generate an offline RL dataset. 
In the trajectory prediction domain, Trajeglish~\cite{philion2024trajeglishtrafficmodelingnexttoken} formulates trajectory prediction as a next token prediction problem using predefined motion tokens. To enhance generalizability, Holigraph Goal~\cite{grimm2025goalbasedtrajectorypredictionimproved} proposes a goal-prediction and goal-based trajectory-prediction approach that leverages the lane structure. While these approaches predict or generate feasible trajectories, they require past trajectories and initial states, which makes it difficult to generate or predict the trajectories.

To further enhance diversity, generating both agent initialization and trajectories has been widely explored.
PathDiffuser~\cite{lee2025pathdiffuserdiffusionmodel} proposes a two-stage diffusion-based approach for traffic scenario generation. For agent initial scene generation, the differential transformer~\cite{ye2024differential} is used to narrow down the attention to the map components. For trajectory generation, candidate trajectories are defined in the Frenet coordinate system while reducing reliance on past motions and improving map compliance in adversarial scenarios.
ScenarioDreamer~\cite{rowe2025scenariodreamer} also proposes a two-stage approach to traffic scenario generation. The scene initialization employs a latent diffusion model that uses latent embeddings captured from a VAE model to represent lane segments and agents' states. Then, the closed-loop simulator from Ctrl-Sim~\cite{rowe2024ctrlsimreactivecontrollabledriving} is leveraged for trajectory generation.
SceneDiffuser~\cite{jiang2024scenediffuserefficientcontrollabledriving} utilizes v-prediction~\cite{salimans2022progressivedistillationfastsampling} to train the diffusion model by using a control mask for controllable generation in both scenario generation and behavior prediction. By using the proposed amortized diffusion, a single denoising function generates future trajectory predictions by refining buffered predictions from previous steps. While these approaches enable the traffic scenario generation, agents' high-level intent is not modeled during initialization, which limits the interpretability of the initial states of the resulting traffic scenario.

Most existing works generate diverse traffic scenarios by factorizing the problem into two stages: generating initial states and then generating trajectories. In contrast, we propose a top-down diffusion-based approach that first generates high-level scenarios, then infills trajectories.

\section{Problem Statements and Preliminaries}
 \label{sec:preliminaries}
This section presents the mathematical notations, the problem formulation, and a brief overview of the diffusion process underlying our approach.

\subsection{Notation}
The vectorized map consists of road elements, including map polygons and map points, whose embeddings are denoted by $\mathcal{M}$. We consider a scene with $N$ agents and a time horizon $H$, with intermediate timestep $h \in \{0,1,\ldots, H\}$. The state of agent $i$ at a given timestep is defined as $a_i = [x_i, y_i, v_i, c_i] \in \mathbb{R}^4$, where $(x_i, y_i) \in \mathbb{R}^2$ denotes position, $v_i \in \mathbb{R}^+$ the speed, and $c_i \in \mathbb{Z}^+$ the agent type. The joint state of all agents in the scene at time step $h$ is $\rvs_h = [a_1, \ldots, a_N] \in \mathbb{R}^{N \times 4}$. The high-level traffic scenarios are denoted by $\rvs_{HL} = [\rvs_0, \rvs_H]$. We denote a positional trajectory of agent $i$ over the time as $\tau_i = [(x,y)_{i,1}, \ldots, (x,y)_{i,H}]$, trajectories for all agents as $\boldsymbol{\tau} = [\tau_1, \ldots, \tau_N]$, and trajectory at time $h$ as $\bm\tau_h$. The real-world distribution is denoted as $q(.)$, and the learned distribution is denoted as $p_\theta(.)$. We denote generated states and predicted trajectories using the hat notation, e.g., $\hat{\bm\tau}$ and $\hat{\rvs}$.

\subsection{Problem Statement}
Given a map $\mathcal{M}$ and the number of agents $N$, our goal is to develop a generative model that approximates the distribution of traffic scenarios from a real-world dataset $q(\boldsymbol\tau | \mathcal{M})$, so that the sampled data from learned distribution $p_\theta(\boldsymbol\tau | \mathcal{M})$ resembles the real-world data. 

Unlike existing approaches, we propose decomposing the problem using the concept of a high-level traffic scenario. We define a high-level traffic scenario $\rvs_{HL} = [\rvs_0, \rvs_H]$, which consists of the initial states $\rvs_0$ and the goal states $\rvs_H$ for all $N$ agents in the map. Using this concept, the problem is decomposed as follows:
\begin{thm}\label{prob:init}
(High-level Traffic Scenario Generation) 
Given a map $\mathcal{M}$ and the number of agents $N$, develop a generative model which enables sampling high-level scenarios ${\rvs_{HL}}$ from $p_\theta(\rvs_{HL} | \mathcal{M})$ that resembles samples from the true distribution $q(\rvs_{HL} | \mathcal{M})$.
\end{thm}

\begin{thm}\label{prob:traj}
(Trajectory Infilling) 
Given a map $\mathcal{M}$ and high-level scenarios ${\rvs_{HL}}$, develop a model for goal-conditioned trajectory prediction that infills trajectories $\boldsymbol{\tau}$ between the sampled initial ${\rvs_0}$ and goal states ${\rvs_H}$.
\end{thm}

\subsection{Diffusion Model}
\label{sec:diffusion_explanation}
Among existing generative models, denoising diffusion probabilistic models (DDPM)~\cite{ho2020denoisingdiffusionprobabilisticmodels} have shown promising results across a variety of domains. DDPM corrupts data with Gaussian noise in a forward pass based on the schedule, and learns to iteratively reverse this process to recover the samples from the original distribution. 

\noindent\textbf{Forward Process.} 
The distribution of the agent's initial and goal states is denoted by $q(\rvs_{HL})$. Over the diffusion steps $T$, Gaussian noise is iteratively added as follows:
\begin{equation*}
 \label{eq:init_forward}
q(\rvs_{HL}^t | \rvs_{HL}^{t-1}, \mathcal{M}) := \mathcal{N}\left(\rvs_{HL}^t; \sqrt{1 - \beta_t} \, \rvs_{HL}^{t-1}, \, \beta_t \, \mathbf{I} \right)
\end{equation*}
where $t\in\{0,1,\ldots, T\}$ is the diffusion steps, $\beta_t$ is the predefined variance schedule at step $t$. Using the Markov property, samples at any step $t$ can be written as:
\begin{equation*}
\label{eq:reverse}
q(\rvs^t_{HL} | \rvs_{HL}^0) = \mathcal{N}\left(\rvs^t_{HL}; \sqrt{\bar{\alpha}_t} \, \rvs_{HL}^0, (1 - \bar{\alpha}_t) \, \mathbf{I} \right)
\end{equation*}
\noindent
in which $\alpha_t = 1 - \beta_t$ represents the variance preserving term and  $\bar{\alpha}_t = \prod_{s=1}^{t} \alpha_s$ controls the noise schedule.

\noindent\textbf{Reverse Process.} 
In the reverse process $p_\theta(\rvs_{HL}^{0:T})$, noisy data is initially drawn from the prior distribution $p_\theta(\rvs_{HL}^T) = \mathcal{N}(\rvs^T_{HL}; \mathbf{0}, \mathbf{I})$. Then, the noisy sample is gradually denoised to recover a noise-free sample, which can be considered as the transition from $\rvs_{HL}^t$ to $\rvs_{HL}^{t-1}$ and expressed as follows:

\begin{equation*}
\label{eqn:reverse}
p_\theta(\rvs_{HL}^{t-1}| \rvs_{HL}^t) = \mathcal{N}(\rvs_{HL}^{t-1}; \bmu_\theta(\rvs_{HL}^t, t), \sigma_t^2 \mathbf{I})
\end{equation*}
where $\sigma_t^2 = \beta_t \frac{1-\bar{\alpha}_{t-1}}{1-\bar{\alpha}_t}$.
Moreover, estimating added noise $\epsilon$ is mathematically equivalent to estimating the clean data $\rvs_{HL}$ but offers better stability.

\noindent\textbf{Guidance Sampling.} 
The reverse process can be guided to steer denoising towards the desired output. While some approaches introduce an additional classifier network to approximate the guidance~\cite{dhariwal2021diffusion}, they require additional training and inference costs. To reduce the computational cost, some works ~\cite{lu2024scenecontrol,wang2024optimizingdiffusionmodelsjoint, jiang2023motiondiffusercontrollablemultiagentmotion} define the guidance cost function $\mathcal{J}(x^0)$ to reflect the desired characteristics, where $x^0$ is the clean data. Then, during the reverse process, the denoised data $\hat{x}^0(x_t)$ is reconstructed by estimating the noise. By taking the gradient $\nabla_{x^t}\mathcal{J}(\hat{x}^0(x_t))$, it is injected to bias the denoising process. For instance, as in SceneControl~\cite{lu2024scenecontrol}, during the reverse process, we can steer the agent's location by biasing with the gradient of the guidance cost function.

In our task, we apply this principle to guide $\rvs_{HL}$ by perturbing the noise prediction as in Eq.~\ref{eqn:guide}. 
\begin{equation}
\label{eqn:guide}
\hat{\bm{\epsilon}}_\theta(
\rvs_{HL}^t, \mathcal{M}) \leftarrow
\bm{\epsilon}_\theta(
\rvs_{HL}^t, \mathcal{M}) - \lambda_{guide}\nabla_{\rvs^t_{HL}}\mathcal{J}(\hat{\rvs}^0_{HL}),
\end{equation}
where $\lambda_{guide}\in \mathbb{R}^+$ is the guidance scale.
\section{\model{}: Approach}
\label{sec:approach}
Building upon the preliminaries and notations in Section~\ref{sec:preliminaries}, we propose a hierarchical diffusion framework for top-down traffic scenario generation.

\subsection{High-Level Traffic Scenario Generation}
\label{sec:hl_gen}
As formulated in Problem~\ref{prob:init}, we first aim to generate the high-level traffic scenarios that resemble the true distribution by jointly modeling the initial and goal states. To learn the real-world distribution, we leverage a DDPM~\cite{ho2020denoisingdiffusionprobabilisticmodels} formulation, and the reverse process $\bm{\epsilon}_{\theta}(\rvs^t_{0,H}, \mathcal{M}, t)$ is parameterized to estimate noise at each diffusion step $t$, given the map $\mathcal{M}$ and the number of agents $N$.

\subsubsection{Architecture}
Based on the DDPM formulation, we now describe the detailed architecture of our framework. As illustrated in Fig.~\ref{fig:overview}A, the high-level scenario generation model consists of a map encoder and a traffic generator.
 
\begin{figure}[htbp]
    \centering
    \includegraphics[clip, trim={0cm 0cm 1.5cm 0cm},width=\linewidth]{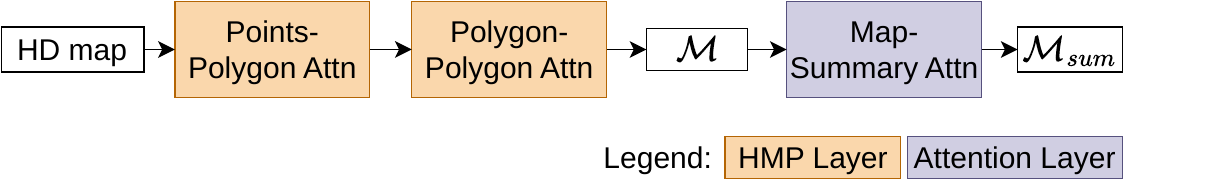}
    \vspace{-7mm}
    \caption{\textbf{Map encoder architecture}. Map polygons are encoded using a heterogeneous message passing (HMP) layer that attends to corresponding map points and surrounding map polygons. Then, the learnable embedding $\mathcal{M}_{sum}$ attends the encoded map polygons to summarize the map structure.}
    \label{fig:mapencoder}
\vspace{-3mm}
\end{figure}

\noindent\textbf{Map Encoder} To efficiently encode the vectorized map structure, we adopt and modify the map encoder from QCNet~\cite{zhou2023query}. Since the goal is to generate the initial states on the map, the map point and map polygon nodes are encoded with their positions and orientations. Then, as illustrated in Fig.~\ref{fig:mapencoder}, map polygon nodes attend map point nodes to learn how the lane is internally structured, and other polygons to learn how they are connected. Moreover, we introduce a map summary layer to summarize the map by attending to encoded map polygons.

\vspace{-2mm}

\noindent\textbf{Traffic Generator} When generating realistic high-level traffic scenarios, the model needs to generate a valid path for each agent while not colliding with other agents. As in the agent initialization model in PathDiffuser~\cite{lee2025pathdiffuserdiffusionmodel}, we adopt the differential transformer~\cite{ye2024differential} to promote the sparse map attention. Unlike PathDiffuser~\cite{lee2025pathdiffuserdiffusionmodel} and SceneControl~\cite{lu2024scenecontrol}, our model attends to map polygon embeddings to reduce the number of attention operations to learn the map structure. 

An overview of the traffic generator is shown in Fig~\ref{fig:overview}A. To encode the high-level scenario, we consider the problem as encoding a sequence of length two, i.e., initial and goal states. Each states are independently encoded with time and agent identity embeddings to maintain temporal and social distinction during the attention operations. The denoising process follows a sequence of attention layers to fuse multi-modal constraints. First, noisy samples attend to a map summary to provide a global spatial context of the given map. This is followed by temporal self-attention to ensure consistency between an agent's own initial and goal states. Then, the sample attends to other agents through a social cross-attention layer to capture the interaction among agents. Lastly, it attends the local map embeddings to condition on the geometry of the given map structure.

\subsubsection{Loss}
 The training objective is to maximize the Variational Lower Bound, but, as shown in~\cite{ho2020denoisingdiffusionprobabilisticmodels}, it can be reduced to predicting noise. By reparameterizing the objective for our high-level scenario $\rvs_{HL}$ generation, the training objective reduces to the following loss function:

\begin{equation}  
\label{eq:hl_obj}
 \mathcal{L}_{\mathbb{HL}}(\theta) =
\mathbb{E} \left[
\left\| {\boldsymbol\epsilon} - \bm{\epsilon}_\theta\left( 
\rvs^t_{HL}, \mathcal{M}, t
\right) \right\|^2
\right]
\end{equation}

To further promote sparser map attention, an additional loss term is introduced to minimize the entropy of the attention weights, as in Eq.~\ref{eq:entropy}. As the differential transformer~\cite{ye2024differential} can output negative attention weights, the attention weights are normalized for entropy computation.
\begin{equation}  
\label{eq:entropy}
\mathcal{L}_{ent}(\theta) =
\mathbb{E} \left[
 -\tilde{w}_i\log(\tilde{w}_i) \right]
\end{equation}
where $\tilde{w}_i = \frac{|w_{i,j}|}{||w_i||_2}$ represents the normalized attention weights of map token $j$ for agent $i$, $w_{i,j}$ denotes the attention weight between agent $i$ and map token $j$, $|.|$ denotes element-wise absolute value, and $||.||_2$ denotes l2-norm of attention weights for agent $i$ across all map tokens.

By combining the Eq.~\ref{eq:hl_obj} and Eq.~\ref{eq:entropy}, the model is trained using the following loss function:
\begin{equation*}  
\label{eq:training_obj}
\mathcal{L}(\theta) =  \mathcal{L}_{\mathbb{HL}}(\theta) +  \lambda_{ent}\mathcal{L}_{ent}(\theta),
\end{equation*}
where $\lambda_{ent} \in \mathbb{R}^+$ is a hyperparameter.

\subsection{Trajectory Infilling}
\label{sec:traj_infill}
In existing trajectory prediction or generation problem formulations, the model must capture how the agent's trajectory rolls out from the initial states or the past trajectories, which requires learning the agent's destination and its behavior along the way. However, with the generated high-level scenarios, the trajectory generation problem can be simplified into a goal-conditioned infilling problem, and the model only needs to know how the agent travels to the goal states.

\subsubsection{Architecture}
\label{sec:infill_arch}
\begin{figure}[htbp]
    \centering
    \includegraphics[width=\linewidth]{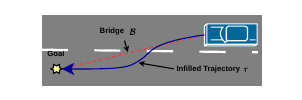}
    
    \vspace{-5mm}
    \caption{\textbf{Illustration of Trajectory Infilling}. Given a ``bridge" (see Sec.~\ref{sec:infill_arch}) between the initial and goal states, a trajectory is produced by refining the predicted distance from the bridge on each time step.}
    \label{fig:infill}
\end{figure}

Given the high-level traffic scenarios, we propose using a line between them as a bridge and designing a model in Fig.~\ref{fig:overview}B to predict the distance from the bridge over time, as illustrated in Fig.~\ref{fig:infill}.

As Algorithm~\ref{alg:infilling_training} describes, to introduce more flexibility to our approach, the infilling model uses a unified architecture that accommodates different constraint settings within a single framework. Specifically, we consider the following cases: (i) hard constraints (HC) on both initial and goal states, (ii) HC on initial states only, (iii) HC on goal states only, and (iv) no HC. Hard constraints set the predicted distance values to zero at the corresponding timestep. Given the initial and goal states, the bridge $\mathcal{B}$ is constructed between them and treated as an anchor. Based on the randomly sampled training mode $m$, the model learns to handle different modes while predicting the distance to the bridge $\mathcal{B}$. After the initial distance estimation, the trajectory is refined by attending to the map and other agents. The weights $\bm\gamma$ are multiplied to apply the constraint depending on the sampled mode. 

\begin{algorithm}[tb]
\caption{Trajectory Infilling}
\label{alg:infilling_training}
\begin{algorithmic}[1]
\renewcommand{\algorithmicrequire}{\textbf{Input:}} 
\Require map $\mathcal{M}$, high-level scenario $\rvs_{HL}$, time horizon $H$, mode $m\sim\{HC_{both}, HC_{init}, HC_{goal}, HC_{none}\}$
\State $\rvh \gets [\frac{i-1}{H-1}]_{i=1}^H$
\Comment{Linearly spaced time vector in $[0, 1]$}

\State $\mathcal{B} \leftarrow (\rvs^0_{H} - \rvs^0_{0})\rvh$
 \Comment{Create evenly spaced bridge}
 
 \State $\bm\gamma \gets \begin{cases} 
\rvh(1-\rvh) & \text{if } m = HC_{both} \\
1-\rvh & \text{if } m = HC_{init} \\
\rvh & \text{if } m = HC_{goal} \\
1 & \text{otherwise}
\end{cases}$\Comment{Mode-based weight}

\State $\hat{\boldsymbol\tau} \leftarrow \mathcal{B} + \bm\gamma f_\phi(\mathcal{M}, \mathcal{B}, m)$
\\
\Return $\hat{\boldsymbol\tau}$
\end{algorithmic}

\end{algorithm}

\begin{figure*}[tp]
    \centering
    \includegraphics[width=\linewidth]{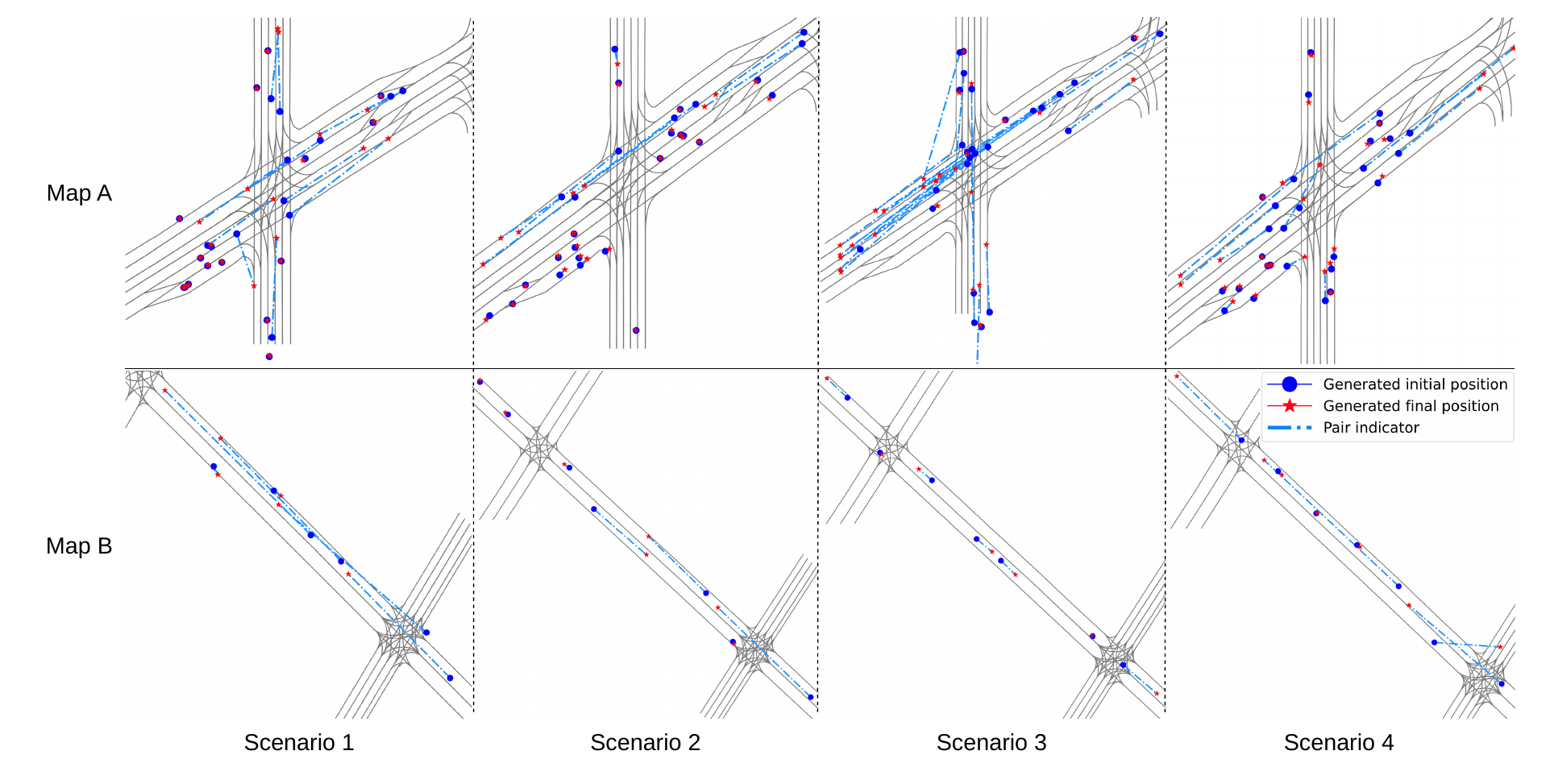}
    \vspace{-10mm}
    \caption{\textbf{Diversity in Generated High-level Traffic Scenario}. Each row contains an identical map, and the number of agents, and each column shows different scenarios. Blue dots indicate the initial positions, and red stars indicate the final positions. The line connecting the two points denotes an initial-goal pair. Generated high-level scenarios show reachable pairs. More visualization is available at \url{https://github.com/CL2-UWaterloo/TrafficDiffuser}.}
    \label{fig:qual_hl_gen}
\vspace{-1mm}
\end{figure*}
\begin{table*}[tb]
\centering
\caption{Performance Comparison of High-level Traffic Scenario Generation}
\vspace{-3mm}
\begin{tabular}{c|c|ccc|cccc}
\hline
                          
                                    \multicolumn{2}{c|}{ }    & \multicolumn{3}{c|}{Common Sense Metrics $\textcolor{red}{\downarrow}$} & \multicolumn{4}{c}{Distributional JSD $\textcolor{red}{\downarrow}$} \\ \cline{3-9} 
                                        
                                    \multicolumn{2}{c|}{ }
                                          & Collision rate (\%)      & Off road rate (\%)      & Near. edge (m)     & Speed            & Lat. Dev.     & Local Density   & Near. Dist.     \\ \hline

\multirow{4}{*}{Initial States}&
GT                                               & 0.50                      & 1.70                     & 1.58               & -                & -             & -               & -               \\
&SC~\cite{lu2024scenecontrol}                     & 4.40                      & 5.25                    & \textbf{1.63}      & 0.16             & 0.212         & 0.177           & 0.092           \\
&PD-init~\cite{lee2025pathdiffuserdiffusionmodel} & \textbf{4.22}            & 5.36                    & 1.64      & 0.1524           & 0.213         & \textbf{0.175}  & \textbf{0.089}  \\
&Ours                                             & 9.80                      & \textbf{5.10}            & 1.81               & \textbf{0.068}   & \textbf{0.200}  & 0.201           & 0.101           \\ \hline

\multirow{2}{*}{Goal States} &GT                                               & 0.50                     & 2.80                    & 1.86               & -                & -             & -               & -               \\
& Ours                                             & 10.6                     & 7.21                    & 2.02               & 0.08             & 0.198         & 0.206           & 0.103           \\ \hline
\end{tabular}
\label{tab:hl_gen_results}
\vspace{-5mm}
\end{table*}
\subsubsection{Loss}
The trajectory infilling model  $f_\phi(.)$ learns to predict the distance from the bridge $\mathcal{B}$, by using the following loss function:
\begin{equation*}  
\label{eq:infill_loss}
\mathcal{L}_{traj}(\phi) = \mathcal{L}_{recon}(\phi) +  \lambda_{s} \mathcal{L}_{smooth}(\hat{\boldsymbol\tau}),
\end{equation*}
where $\lambda_s$ is a hyperparameter to balance temporal smoothing and accuracy. The reconstruction loss is defined as
$\mathcal{L}_{recon}(\phi) = ||\hat{\boldsymbol{\tau}}- \boldsymbol\tau||^2$, where the predicted trajectory is reconstructed as $\hat{\boldsymbol{\tau}} = \mathcal{B} + \bm\gamma f_\phi(\mathcal{M}, \mathcal{B}, m)$. To avoid abrupt motions, a smoothness penalty is introduced as$\mathcal{L}_{smooth}(\hat{\boldsymbol\tau}) =\sum_{h=0}^{H}||\hat{\boldsymbol\tau}_{h+1} - \hat{\boldsymbol\tau}_h||_2$, which computes the mean squared first-order difference over the trajectory horizon.
 
\section{Experiments}
\label{sec:experiments}

In this section, we demonstrate the generation of high-level traffic scenarios. We show how the jointly generated initial and goal states improve the interpretability by providing more context for each agent's initial placement. Also, we show how the generated high-level traffic scenarios can be leveraged for trajectory generation, demonstrating their compatibility through experiments.

\subsection{Experimental Setup}
\subsubsection{Dataset}
Argoverse 2 Motion Forecasting Dataset~\cite{wilson2023argoverse2generationdatasets}, which is a large-scale real-world dataset for motion planning and prediction, is used to capture the distribution of real-world traffic scenarios. As explained in Section~\ref{sec:preliminaries}, for diffusion models, unit-variance Gaussian noise is added based on the predefined schedule during the forward process. For training, the input data should be scaled to have a variance comparable to that of the noise, ensuring meaningful signal-to-noise ratios. Therefore, during data preprocessing, the speed and positions of the agents and map components are z-normalized using the mean and standard deviation of the agents' initial and final states. We consider 5 agent categories: vehicle, bus, pedestrian, bicycle, and motorcycle.

\subsection{High-level Traffic Scenario}

\subsubsection{Baselines}
Although we generate both initial and goal states of the agents in the scene, we compare our initialization results against the most relevant agent initialization approaches in the literature. Specifically, we evaluate against SceneControl(SC)~\cite{lu2024scenecontrol}  replica, as it is not publicly available, and PathDiffuser's agent initialization (PD-Init)~\cite{lee2025pathdiffuserdiffusionmodel}.

\subsubsection{Performance Measure}
The generated high-level traffic scenarios contain two time steps, and each time step is evaluated separately. As in~\cite{lu2024scenecontrol, lee2025pathdiffuserdiffusionmodel, rowe2025scenario}, we assess the realism of the generated scenarios using Common Sense and Jensen-Shannon Divergence (JSD). 
Common Sense includes (i) \textbf{Collision Rate}, which measures the ratio of collided agents, (ii) \textbf{Off-road Rate}, which measures the ratio of off-road agents, and (iii) the distance to the nearest lane center point(\textbf{Near. Edge}). To measure the distributional distance between ground truth and generated output, the distributions are estimated using histograms. Then, using JSD, the distance is measured for (i) \textbf{Speed}, (ii) distance to the nearest lane center (\textbf{Lat. Dev.}), (iii) distance to 5 nearest agents (\textbf{Local Density}), and (iv) distance to the nearest agent (\textbf{Near. Dist.}).

\subsubsection{Quantitative Results}
Table~\ref{tab:hl_gen_results} shows the evaluation results for initial and goal states in separate rows. Although the task became more complex by modeling the high-level traffic scenario, we observe that the off-road rate is reduced by 2.8\% compared to the PD-init. In terms of JSD, speed distribution is closer to the ground truth by 55.3\%, compared to the speed output distribution of SC. Also, we observe that \model{} shows an increase in the collision rate and the distance to the nearest edge, which is expected as we use encodings of map polygons instead of map points. For fair comparison, the results reported in Table~\ref{tab:hl_gen_results} were obtained without the use of guidance sampling. As shown in SceneControl~\cite{lu2024scenecontrol}, guidance sampling can improve realism, which improves quantitative and qualitative results. 
\subsubsection{Qualitative Results}
Generated high-level traffic scenarios are visualized in Fig.~\ref{fig:qual_hl_gen}. We observe that the generated high-level scenario consists of pairs of points, the initial and goal states of the same agent, with reachable paths on the map. To show the diversity of generated scenarios, we generate multiple scenarios from the identical map. Each row in Fig.~\ref{fig:qual_hl_gen} shares the same map, while each column displays variations with a different number of agents. To leverage the benefit of diffusion models, guidance sampling, explained in Sec.~\ref{sec:diffusion_explanation}, is used to bias the reverse process. As guidance functions, the distances to neighboring agents and to the lane center are used.

\subsubsection{Interpretability}
One of the main contributions of the proposed framework is that the generated initial states are better interpretable. While baseline approaches only generate agent initial states, our approach provides agents' initial states with their intent by indicating where the vehicle is trying to reach. For instance, as shown in Fig.~\ref{fig:qual_hl_gen}, each agent is explicitly paired with a corresponding goal state. By comparing the initial and goal states, we can infer the vehicle's intention. A close distance between them indicates the vehicle is stationary, whereas a larger distance between them indicates a moving vehicle. Compared to existing approaches, initial-goal pairing enhances scenario interpretability by showing the underlying intent of each agent and provides a preview of the traffic scenario.

\subsection{Trajectory Infilling}
Given the map and high-level traffic scenarios, trajectory generation can be simplified to a goal-conditioned trajectory prediction by connecting the initial and end states.

\subsubsection{Metrics}
As the trajectories are infilled by fixing the initial and goal states, we evaluate the trajectories by comparing with the groundtruth using standard metric for motion prediction~\cite{rowe2024ctrlsimreactivecontrollabledriving,grimm2025goalbasedtrajectorypredictionimproved}, which are (1) Average Displacement Error (\textbf{ADE}) that measures average displacement across the prediction horizon $H$, (2) Final Displacement Error (\textbf{FDE}) that measures the displacement error of the final states, and (3) Miss Rate (\textbf{MR}) that evaluates the rate of agents with FDE is above 2 meters.

\subsubsection{Quantitative Results}
As shown in Table~\ref{tab:traj_inifll_results}, MR is all zero, and ADE is close to groundtruth, which is expected since we use the bridge $\mathcal{B}$ between initial and goal position as an anchor. Also, as the weight $\bm\gamma$ forces the distance to zero under the given constraints, we observe that FDE is zero if the goal states are fixed. 

\subsubsection{Qualitative Results}
\begin{figure}[htbp]
    \centering
    \includegraphics[width=\linewidth]{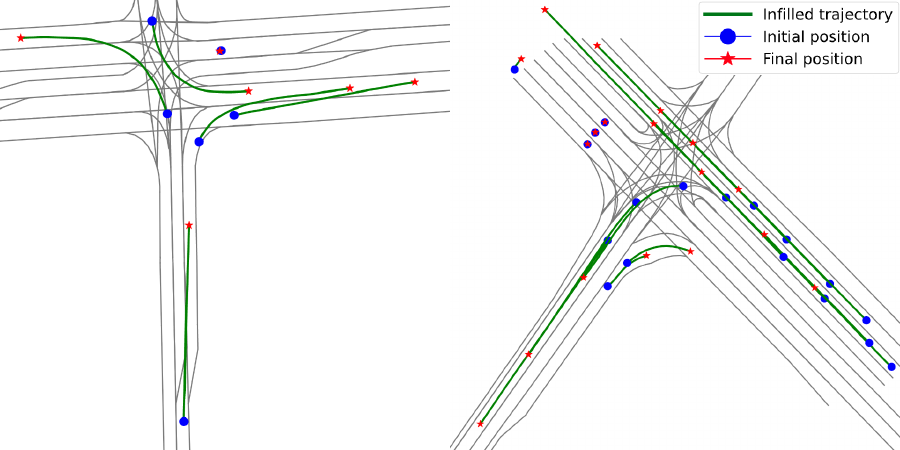}
    \vspace{-5mm}
    \caption{\textbf{Visualization of Trajectory Infilling}. Given the ground-truth initial and goal states, the model infills the states by connecting them with a kinematically feasible trajectory.}
    \label{fig:qual_infill}
\vspace{-6mm}
\end{figure}

Fig.~\ref{fig:qual_infill} shows infilled trajectories between the high-level scenarios. We can observe that, given high-level scenarios, the model produces trajectories between the initial and goal states while following the map structure, even when changing lanes or turning left.

\begin{table}[tb]
\centering
\caption{Quantitative Evaluation of Trajectory Infilling}
\vspace{-3mm}
\begin{tabular}{c|c|ccc}
\hline
            Model                     &                 Hard Constraints                                 & \begin{tabular}[c]{@{}c@{}}ADE \\ (m)$\textcolor{red}{\downarrow}$\end{tabular} & \begin{tabular}[c]{@{}c@{}}FDE \\ (m) $\textcolor{red}{\downarrow}$\end{tabular} & \begin{tabular}[c]{@{}c@{}}MR\\ (\%) $\textcolor{red}{\downarrow}$\end{tabular} \\ \hline
\multirow{4}{*}{Ours} & Initial and Goal States                              & 0.52                                                                           & 0                                                                               & 0                                                                               \\
                                 & Initial States                             & 0.53                                                                           & 0.16                                                                            & 0.19                                                                               \\
                                 & Goal States                                & 0.53                                                                           & 0                                                                             & 0                                                                               \\
                                 & None                                            & 0.53                                                                           & 0.1                                                                            & 0.04                                                                               \\ \hline
     PD-init~\cite{lee2025pathdiffuserdiffusionmodel}              & Initial States&      1.86                                                                    &                  7.16                                                                & 43.84                                                                                \\ \hline

\end{tabular}
\label{tab:traj_inifll_results}
\vspace{-5mm}
\end{table}

\begin{figure*}[t!]

    \centering
    \includegraphics[width=0.9\linewidth]{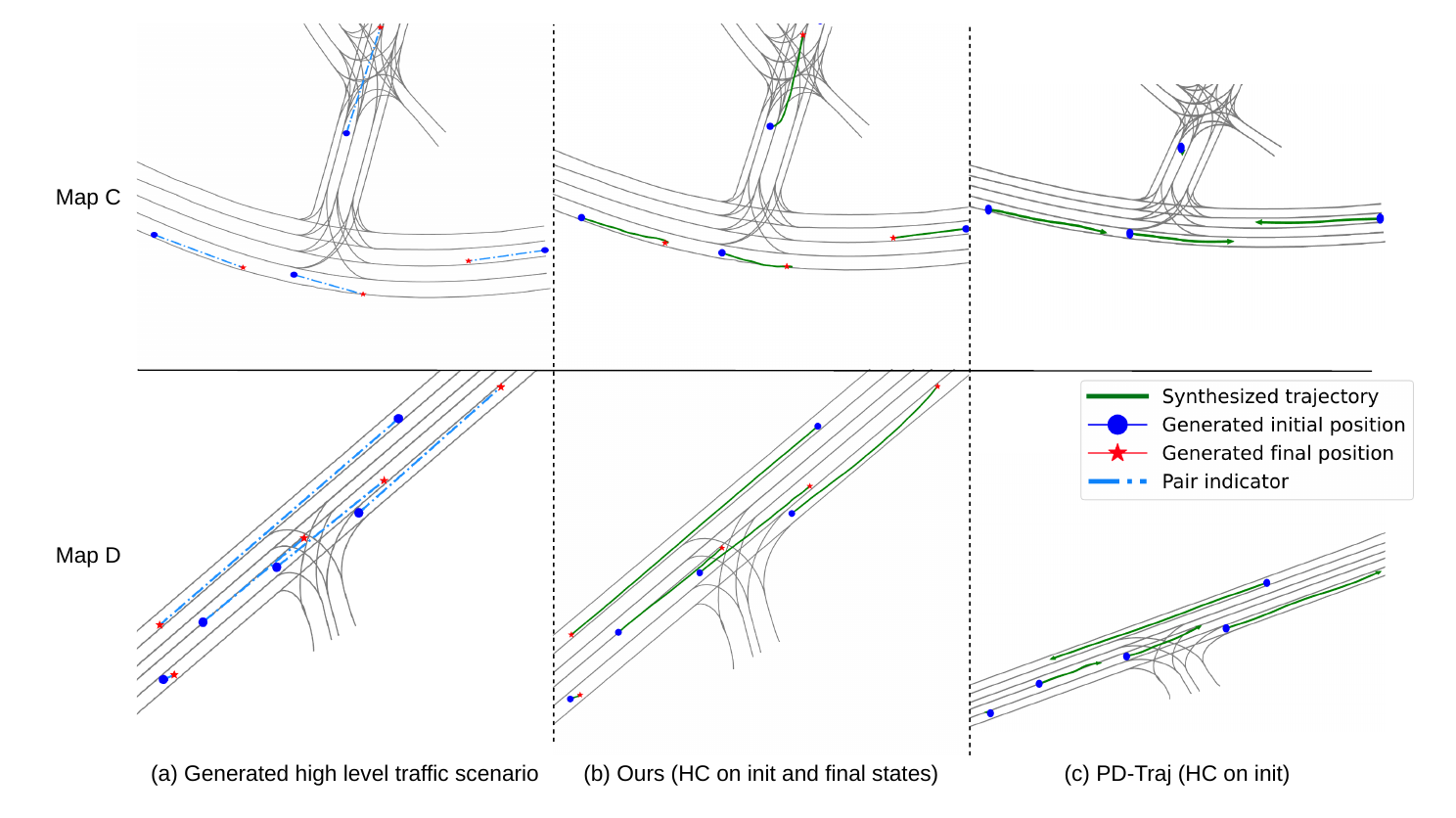}
    \vspace{-5mm}
    \caption{\textbf{Visualization of End-to-end Traffic Scenario Generation}. Each row has an identical map structure to demonstrate trajectory generation given a high-level scenario. (a) visualizes the sampled high-level traffic scenarios containing the initial and goal states, and (b) illustrates the infilled trajectory between them with hard constraints (HC). (c) demonstrates the compatibility of the framework with the PD-Traj model to generate trajectories conditioned on the given initial states.}
    \label{fig:end2end}
\vspace{-7mm}
\end{figure*}

\subsection{End-to-end Process for Complete Scenario Generation}
We now evaluate the entire pipeline, in which the generated multi-agent high-level traffic scenarios are used to produce trajectories. As illustrated in Fig.~\ref{fig:end2end}(a), the high-level scenario consists of synthesized pairs of agents. Then, Fig.~\ref{fig:end2end}(b) shows the predicted trajectory using our trajectory infilling model.

\noindent\textbf{Compatibility with Existing Trajectory Generator}
To demonstrate how the generated high-level scenarios can be used, we integrate them with the trajectory generation model from PathDiffuser (PD-Traj)~\cite{lee2025pathdiffuserdiffusionmodel}. As Fig.~\ref{fig:end2end}(c) shows, our high-level traffic scenario generator can be combined with an existing trajectory generation model to promote diverse traffic scenario generation.

\section{Discussion}
\label{sec:conclusion}
\noindent{\textbf{Summary.}}
We presented \model{}, a framework that factorizes the multi-agent traffic scenario generation into a high-level traffic scenario generation and a goal-conditioned trajectory infilling. By jointly modeling the initial and goal states, our approach captures reachable initial and goal states, as well as social interactions, from a real-world driving dataset. This hierarchical structure provides enhanced interpretability from high-level traffic scenarios and simplifies the trajectory generation task into a goal-conditioned prediction problem.
Our framework not only optimizes the generative process but also maintains modular compatibility with existing models in the trajectory generation domain.

\noindent\textbf{Limitations and Future Work.} Despite these contributions, our approach has certain limitations.
In the proposed framework, stochasticity is confined to the high-level scenario generation, while the trajectory generation is formulated as a deterministic, goal-conditioned infilling task rather than a probabilistic process. While this accelerates inference, it limits the diversity of trajectories for a given high-level scenario. To address this, we plan to integrate generative models into trajectory infilling to better capture the multimodality of agent behavior and extend it to conditional generation based on partially observed agent trajectories.

\noindent\textbf{Conclusion.} \model{} demonstrates that modeling traffic scene initialization with agents' intent enhances the interpretability of synthesized traffic scenarios. By factorizing traffic scenario generation into high-level scenario generation and goal-conditioned trajectory infilling problems, we provide a modular and efficient pipeline for synthesizing complex multi-agent traffic scenarios. Ultimately, this framework provides an efficient, interpretable foundation for developing and validating autonomous systems in realistic traffic scenarios.
\vspace{-2mm}
	\bibliographystyle{IEEEtran}
	\bibliography{root}
	
\end{document}